\documentclass[10pt,letterpaper]{article}
\usepackage[margin=0.9in]{geometry}
\usepackage[T1]{fontenc}
\usepackage{newtxtext,newtxmath}
\usepackage[font=small,labelfont=bf,skip=6pt]{caption}
\usepackage{microtype,amsmath,booktabs,array,tabularx,xcolor,tikz,pgfplots,enumitem}
\usepackage[numbers,sort&compress]{natbib}
\usepackage[colorlinks=true,urlcolor=linkblue,citecolor=linkblue,linkcolor=linkblue]{hyperref}
\usetikzlibrary{arrows.meta,positioning,calc,shapes.geometric}
\pgfplotsset{compat=1.18}

\definecolor{ink}{HTML}{1F2933}
\definecolor{muted}{HTML}{5B6670}
\definecolor{linkblue}{HTML}{1F4E8C}
\definecolor{cAd}{HTML}{2A6FB0}      
\definecolor{cFx}{HTML}{C8711F}      
\definecolor{cPl}{HTML}{8A5FB0}      
\definecolor{cOb}{HTML}{8C939B}      
\definecolor{cIv}{HTML}{7A6A4F}      
\definecolor{labbg}{HTML}{F1F5FA}
\definecolor{tgtbg}{HTML}{F7F4EE}
\definecolor{bid}{HTML}{3C8D8A}
\definecolor{ask}{HTML}{C0625A}
\tikzset{
  >=Latex,
  every picture/.style={line width=.6pt},
  box/.style={draw=ink!70,fill=white,rounded corners=1.5pt,align=center,
              inner sep=3.5pt,minimum height=.75cm,font=\small},
  flow/.style={->,ink!75,line width=.7pt},
  damp/.style={-{Bar[width=6pt]},line width=1pt},
  lbl/.style={font=\scriptsize,text=muted,align=center},
}
\pgfplotsset{
  msb/.style={
    axis line style={ink!60},tick style={ink!60},
    major tick length=2.5pt,
    tick label style={font=\scriptsize,text=ink},
    label style={font=\small,text=ink},
    every axis title/.style={font=\small\bfseries,at={(0,1)},anchor=south west,yshift=4pt},
  },
}

\newcommand{\R}{\mathcal{R}}
\newcommand{\act}[1]{\texttt{\small #1}}
\newcommand{\Obs}{\textsc{Observe-only}}
\newcommand{\Ad}{\textsc{Adaptive-lab}}
\newcommand{\Fx}{\textsc{Fixed-mix-lab}}
\newcommand{\Pl}{\textsc{Placebo-tools}}
\newcommand{\ObsS}{\textsc{Obs}}
\newcommand{\AdS}{\textsc{Adapt}}
\newcommand{\FxS}{\textsc{Fixed}}
\newcommand{\PlS}{\textsc{Placebo}}

\hypersetup{pdftitle={CrashGym: Measuring the Decision Value of Experimentation for LLM Agents in Counterfactual Market Crashes},pdfauthor={Haocheng Lu, Haoxuan Cheng, Mingyang Chen, Minjun Zhu}}

\begin{document}

\begin{center}
{\LARGE\bfseries CrashGym: Measuring the Decision Value of\\[3pt] Experimentation for LLM Agents\\[3pt] in Counterfactual Market Crashes\par}
\vspace{12pt}
{\large Haocheng Lu$^{1,*,\dagger}$ \qquad Haoxuan Cheng\textsuperscript{2,*} \qquad Mingyang Chen\textsuperscript{3} \qquad Minjun Zhu\textsuperscript{4}}\par
\vspace{6pt}
{\small \textsuperscript{1}Columbia University \qquad \textsuperscript{2}Cornell University\par
\textsuperscript{3}Hithink RoyalFlush Information Network Co., Ltd. \qquad \textsuperscript{4}New York University\par}
\vspace{3pt}
{\small \textsuperscript{*}Equal contribution. \qquad $^{\dagger}$Corresponding author: \href{mailto:haocheng409@gmail.com}{\texttt{haocheng409@gmail.com}}\par}
\end{center}
\vspace{4pt}

\begin{abstract}
\noindent
Language-model agents are increasingly asked to run experiments, yet benchmarks rarely measure whether those experiments improve the decisions they are meant to inform. We introduce \emph{CrashGym}, a deterministic, replayable limit-order-book simulator in which an agent must choose between a deleveraging intervention and a liquidity backstop for a market about to suffer a shock-triggered crash. Before committing, the agent may query intervention experiments in a laboratory realization of the market. Its choice is then scored on an independently seeded target realization in which the simulator executes \emph{every} feasible intervention, so regret relative to the best intervention is computed by directly simulating every alternative on the same target repeats, with no human or model-based judging. The simulator implements each amplification channel---persistent price impact, liquidity withdrawal, margin-driven liquidation, and an evolving strategy mix---as an explicit mechanism that can be switched on in stages. In a prospectively specified study on 24 newly sampled worlds, laboratory access reduces the mean regret of the one deployment with a complete record (Luna) from 8.878 to 3.785, a paired reduction of 5.092 utility units (nominal 95\% world-bootstrap interval: 1.028--9.507), or 57.4\% of the \Obs{} mean. Prospective comparisons for two further deployments are incomplete; in a separate exploratory study, mean regret is lower with laboratory access for all four deployments evaluated. A laboratory whose strategy mix evolves endogenously yields no established advantage over one with a fixed mix. CrashGym thus measures the value of experimentation as avoided decision loss and lets simulator detail be judged by the decisions it supports.
\end{abstract}

\section{Introduction}
\label{sec:intro}

A market is about to fall sharply. A policy agent must recommend one of two interventions: force leveraged holders to cut their positions, or post a standing bid that supplies liquidity. Both are defensible \emph{a priori}. Deleveraging can pre-empt the cycle of margin calls and forced selling, whereas a liquidity backstop can absorb that selling before it moves prices~\citep{brunnermeier2009,geanakoplos2010,shleifer2011}. Which intervention is better depends on which amplification channel dominates in the market at hand. We ask whether an agent that can first experiment on a laboratory copy of the market chooses better, and by how much.

Language-model agents are increasingly evaluated as experimenters: they design studies, run code, and report findings~\citep{lu2024aiscientist,huang2024mlagentbench,jansen2024discoveryworld}. Most evaluations, however, score the research process or its products---task completion, explanation quality, or recovery of a latent structure. Completing an experiment and \emph{benefiting} from it are different outcomes. Measuring the benefit requires two ingredients that are rarely available together: the same decision-maker observed with and without experimental access, and the counterfactual outcome of every action it could have taken. The second ingredient is unobservable in historical data.

\begin{figure}[t]
\centering
\begin{tikzpicture}[x=1cm,y=1cm]
\fill[labbg,rounded corners=4pt] (0,0) rectangle (7.6,4.75);
\fill[tgtbg,rounded corners=4pt] (8.9,0) rectangle (16.5,4.75);
\node[anchor=north west,font=\small\bfseries,text=ink] at (.15,4.65) {(a) Laboratory realization};
\node[anchor=south west,font=\scriptsize,text=muted] at (.15,.08) {seed cohort $j$ \,$\cdot$\, 8 repeats};
\node[anchor=north west,font=\small\bfseries,text=ink] at (9.05,4.65) {(b) Independent target realization};
\node[anchor=south east,font=\scriptsize,text=muted] at (16.35,.08) {seed cohort $1-j$ \,$\cdot$\, 8 repeats};
\node[box,minimum width=3.6cm] (world) at (8.25,5.7) {World $w$: shared constructor\\[-1pt]{\scriptsize\color{muted}parameters $\cdot$ participants $\cdot$ shock}};
\draw[flow,ink!45] (world.west) -| (3.8,4.75);
\draw[flow,ink!45] (world.east) -| (12.7,4.75);
\draw[ink!45,rounded corners=3pt,fill=white!60!labbg] (.35,.65) rectangle (2.95,3.6);
\node[lbl,anchor=south] at (1.65,3.6) {experiment menu};
\node[box,minimum width=2.2cm,font=\small\ttfamily] (p1) at (1.65,3.05) {LEVERAGE};
\node[box,minimum width=2.2cm,font=\small\ttfamily] (p2) at (1.65,2.12) {LIQUIDITY};
\node[box,minimum width=2.2cm,font=\small\ttfamily] (p3) at (1.65,1.2) {NULL};
\node[box,minimum width=2.0cm,minimum height=1.3cm,draw=cAd,line width=.9pt] (agent) at (6.0,2.12) {\textbf{LLM agent}\\[-1pt]{\scriptsize\color{muted}$\le 8$ calls}};
\draw[flow] ($(agent.west)+(0,.32)$) -- node[lbl,above]{\texttt{run\_experiment}} (2.95,2.44);
\draw[flow] (2.95,1.8) -- node[lbl,below]{8-repeat summary} ($(agent.west)+(0,-.32)$);
\node[box,minimum width=1.75cm,minimum height=1.3cm,draw=cAd,line width=.9pt] (act) at (10.1,2.12) {Committed\\[-1pt]action $a$};
\draw[->,cAd,line width=1.1pt] (agent.east) -- node[font=\tiny,text=cAd,above]{\texttt{submit\_decision}} (act.west);
\draw[ink!40,dashed,rounded corners=3pt] (11.4,.8) rectangle (13.8,3.55);
\node[lbl,anchor=south] at (12.6,3.55) {never shown to agent};
\node[box,minimum width=2.0cm] (ucap) at (12.6,3.0) {$\bar U(\act{CAP})$};
\node[box,minimum width=2.0cm] (ubs) at (12.6,1.35) {$\bar U(\act{BACKSTOP})$};
\node[lbl] at (12.6,2.17) {every feasible\\[-2pt]action executed};
\node[box,minimum width=1.85cm,minimum height=1.3cm] (reg) at (15.4,2.12) {\textbf{Regret}\\[-1pt]{\scriptsize$\max_{a'}\bar U(a')$}\\[-3pt]{\scriptsize$-\,\bar U(a)$}};
\draw[flow] (ucap.east) -- (14.1,3.0) |- ($(reg.west)+(0,.25)$);
\draw[flow] (ubs.east) -- (14.1,1.35) |- ($(reg.west)+(0,-.25)$);
\draw[flow] (act.south) -- (10.1,.5) -| node[lbl,below,pos=.2]{chosen $a$} (reg.south);
\end{tikzpicture}
\caption{\textbf{Task overview.} A world constructor $w$ is instantiated with two disjoint seed cohorts. (a)~In the laboratory cohort the agent may run intervention experiments and receives public eight-repeat summaries. (b)~Its committed action is scored on the other cohort, where the simulator executes both feasible interventions; regret is the utility gap to the better one. Target outcomes are never visible to the agent. Under \Obs{}, the agent decides from the initial observation alone.}
\label{fig:workflow}
\end{figure}
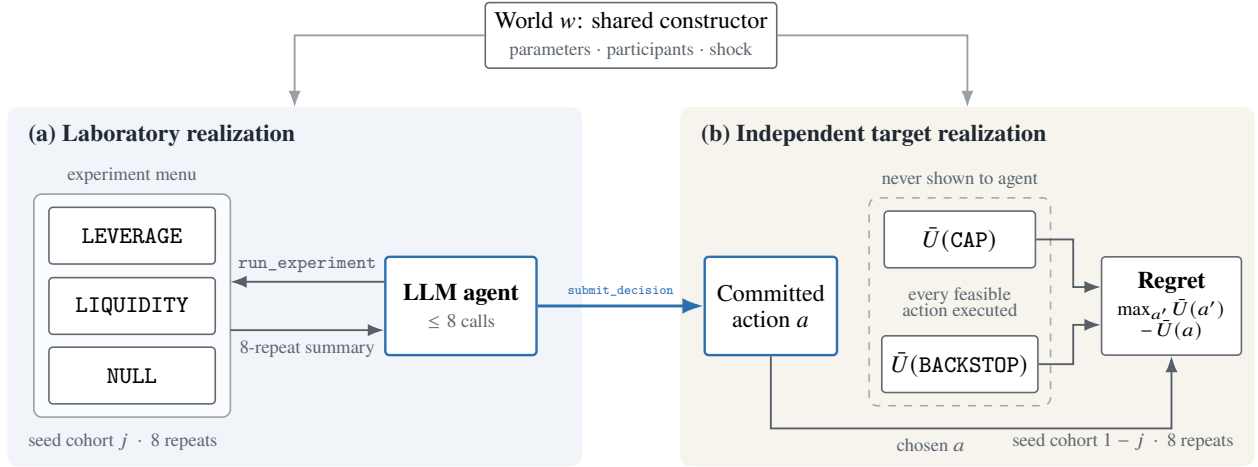

\textbf{Approach.} CrashGym supplies both ingredients inside a deterministic agent-based market (Figure~\ref{fig:workflow}). An agent receives a public observation, may query intervention experiments in a laboratory realization, and then commits to one intervention. The simulator executes \emph{every} feasible intervention on an independent target realization that the agent never observes. Regret---the utility gap between the best intervention and the chosen one---is therefore exact for the realized target repeats. Comparing regret across information conditions yields the gross decision value of experimentation, a realized counterpart of the classical value of sample information~\citep{lindley1956,howard1966,raiffa1961}. Because the simulator's amplification channels are explicit mechanisms that can be enabled in stages, the same outcome also tests whether added simulator detail improves the decision it is meant to support.

\textbf{Contributions.}
\begin{itemize}[leftmargin=1.4em,itemsep=1pt,topsep=2pt]
\item \textbf{A counterfactual market laboratory.} We describe an exchange core with exact integer accounting, event-sourced replay, and counter-based randomness, together with a crash construction whose amplification channels can be enabled in cumulative stages (Section~\ref{sec:sim}).
\item \textbf{A decision-grounded evaluation task.} A laboratory in which the agent acquires interventional evidence is separated from an independently seeded target on which every feasible intervention is executed and scored. This yields exact counterfactual regret for every decision (Section~\ref{sec:task}).
\item \textbf{Controlled evidence.} With identical targets, we compare no access, two laboratory designs, and a tool placebo, sampling new worlds prospectively and using a stratified whole-world bootstrap (Section~\ref{sec:protocol}). Laboratory access lowers Luna's mean regret by 5.092 units on 24 newly sampled worlds, meeting the prespecified criterion. An evolving strategy mix in the laboratory yields no established additional benefit (Section~\ref{sec:results}).
\end{itemize}
The measured value is avoided loss in simulator utility units. Net value would additionally require the costs of experimentation and inference, which we do not measure.

\section{Related work}
\label{sec:related}

\textbf{LLM agents as experimenters.} Tool-using language models interleave reasoning with external actions~\citep{yao2023react,schick2023toolformer}, and a growing set of environments evaluates them as scientists. ScienceWorld and DiscoveryWorld assess task completion and discovery in simulated worlds~\citep{wang2022scienceworld,jansen2024discoveryworld}; MLAgentBench evaluates machine-learning experimentation~\citep{huang2024mlagentbench}; and end-to-end systems generate complete research papers~\citep{lu2024aiscientist}. CausaLab distinguishes predictive success from recovery of causal structure, and CausalGame evaluates causal reasoning and experimental protocols in games~\citep{yang2026causalab,chen2026causalgame}; see also \citet{kiciman2023causal}. We measure a different quantity: how much an agent's experiments improve a subsequent, independently scored decision. An agent may use informative intervention results well without reconstructing the latent mechanism.

\textbf{LLM agents in finance.} Domain-adapted models and agents have been applied to financial analysis and trading~\citep{wu2023bloomberggpt,yu2023finmem,xiao2024tradingagents}, and InvestorBench evaluates financial decision-making across tasks~\citep{li2025investorbench}. These evaluations typically rely on historical data, in which the outcomes of untaken actions are unavailable. Our simulator supplies those counterfactuals for a policy intervention, at the cost of using synthetic rather than historical markets.

\textbf{Market simulation.} Agent-based models have long been used to study market dynamics~\citep{gode1993,lebaron2006,farmer2009}. ABIDES supports high-fidelity, message-level market research~\citep{byrd2019abides}; JAX-LOB targets accelerated, parallel limit-order-book simulation~\citep{frey2023jaxlob}; and realism metrics assess how closely simulated order books reproduce stylized facts~\citep{vyetrenko2020}. Our crash construction draws on the leverage-cycle and fire-sale literature, in which margin constraints, forced selling, and liquidity withdrawal reinforce one another~\citep{brunnermeier2009,geanakoplos2010,thurner2012,shleifer2011}. Rather than maximizing realism, we use the simulator as an instrument that supplies exact counterfactual intervention outcomes and lets each amplification channel be switched on or off.

\textbf{Value of information and decision-aware model evaluation.} Statistical decision theory values an experiment by the improvement in the expected utility of the decision it informs~\citep{lindley1956,howard1966,raiffa1961,chaloner1995}. Our gross value $V_b$ (Section~\ref{sec:discussion}) is a realized, empirical counterpart for an LLM agent. A related literature argues that models should be judged by the decisions they induce rather than by predictive accuracy alone~\citep{donti2017,elmachtoub2022,lambert2020mismatch}, and the value-equivalence principle motivates retaining only those model distinctions that matter for planning~\citep{grimm2020value}. Our comparison of laboratory designs examines this motivation empirically; it does not test Bellman-operator equivalence. The setting also relates to sim-to-real transfer, in which the deployment environment differs from the training simulator~\citep{tobin2017domain}.

\section{The CrashGym simulator}
\label{sec:sim}

The simulator is a discrete-time, single-asset market with a central limit-order book. We describe its components in turn: the exchange core, the participants, the episode schedule, the amplification mechanisms, the evolving strategy mix, the mechanism stages, and the determinism and replay layer. Parameter values refer to the crash construction used in all experiments; Appendix~\ref{app:params} lists them in full.

\subsection{Exchange core}
\textbf{Order book and matching.} Prices are integers in ticks. Each side of the book maps a price level to a first-in-first-out queue, giving strict price--time priority. The only order primitive is a limit order, together with cancellation by the owning account. A marketable order is emulated as a limit order priced 100 ticks through the best opposite quote; any unfilled remainder is cancelled immediately, so it behaves as an immediate-or-cancel order with a price band. Incoming orders walk the opposite side best-first, and each fill executes at the \emph{resting} order's price for the smaller of the two quantities; a partially filled resting order keeps its queue position. The reference price $m_t$ is the sum of best bid and best ask, i.e.\ the midpoint in half-tick units, so it remains an exact integer. If one side of the book is empty, the reference falls back to the last two-sided quote, and aggressive orders are priced from the last published bid or ask. Book invariants---no empty levels, no zero quantities, strictly increasing priorities, and no crossed book---are checked after every mutation.

\textbf{Accounts, risk admission, and fees.} Each participant owns an account with integer cash and inventory, with positions tracked as first-in-first-out lots. Before an order is accepted, the exchange checks a position limit ($10^6$ lots) and a leverage limit: the worst-case position notional along the order's potential fill path may not exceed $20\times$ the account's initial cash. Rejected orders are logged and have no effect. A fee of one cash unit per lot is charged to \emph{both} counterparties of every fill; resting orders that never fill pay nothing. Every fill must conserve total inventory and total cash plus fees, global invariants are re-checked after each operation, and a full canonical audit closes every run.

\subsection{Participants}
Table~\ref{tab:participants} summarizes the six participant classes and the two exogenous accounts. Each step, a \emph{market maker} posts a 40-level ladder on each side, at offsets of $2+k$ ticks ($k=0,\dots,39$) around a center that tracks the previous reference price. Its depth per level is a world-specific $\bar Q\in\{24,\dots,56\}$ lots in the initial ladder (step~0), 400 lots during steps 1--7, and $\bar Q$ again from the shock onward. Every world therefore experiences a scheduled liquidity cliff at the shock, in addition to the stress-driven withdrawal described below. Ladders are reconciled rather than rebuilt: unchanged levels keep their queue priority, and superseded orders are cancelled. \emph{Noise traders} submit eight marketable orders of four lots per step, each a buy or sell with probability~$1/2$. A \emph{fundamental trader} trades four lots per tick of deviation between the fundamental value $v$ and the midpoint. A \emph{trend follower} trades $4e_t$ lots, where $e_t=\operatorname{trunc}\bigl((3e_{t-1}+r_{t-1})/4\bigr)$ is a moving average of past returns $r_t=m_t-m_{t-1}$ (half-ticks) and $\operatorname{trunc}$ rounds toward zero. The fundamental value is held constant at $v=10{,}000$ ticks: no news arrives, so post-shock dynamics are responses to the shock, the scheduled post-shock depth, and the intervention. The \emph{leveraged trader} builds a long position at step~1 and afterwards trades only through forced liquidation or the deleveraging intervention. The \emph{adaptive participant} chooses each step between two candidate strategies (Section~\ref{sec:ecology}).

\begin{table}[t]
\centering\small
\begin{tabularx}{\linewidth}{@{}l>{\raggedright\arraybackslash}Xl@{}}
\toprule
Participant & Rule per step & Active from\\
\midrule
Market maker & 40-level ladder per side at $\pm(2+k)$ ticks; $\bar Q$ lots/level at step 0, 400 during steps 1--7, $\bar Q$ from the shock on, thinned under stress (Eq.~\ref{eq:withdrawal}) & all configurations\\
Noise traders & 8 marketable orders of 4 lots, side drawn with probability $1/2$ & all configurations\\
Leveraged trader & buys 160--240 lots at step 1; forced sales when below maintenance margin (Eq.~\ref{eq:margin}) & margin rule: \emph{leverage} stage\\
Fundamental trader & 4 lots per tick of $v-\text{mid}$, marketable & \emph{leverage} stage\\
Trend follower & $4e_t$ lots, $e_t=\operatorname{trunc}((3e_{t-1}+r_{t-1})/4)$, marketable & \emph{leverage} stage\\
Adaptive participant & after the shock, draws the \textsc{Fundamental} or \textsc{Trend} candidate from shares $(s_{\mathrm F},s_{\mathrm{Tr}})$; shares fixed or evolving (Eq.~\ref{eq:replicator}) & \emph{leverage} stage\\
\midrule
Shock account (exogenous) & marketable sale of 160--240 lots at step 8 & all configurations\\
Intervention account & executes the chosen intervention (Section~\ref{sec:interventions}) & all configurations\\
\bottomrule
\end{tabularx}
\caption{\textbf{Participants in the crash construction.} Ranges are sampled per world (Appendix~\ref{app:params}). ``Active from'' refers to the mechanism stages of Figure~\ref{fig:mechanisms}a; the \emph{leverage} stage is the first stage used in our laboratories.}
\label{tab:participants}
\end{table}

\subsection{Episode schedule}
An episode lasts $T=96$ steps (Figure~\ref{fig:episode}a). At step~0 the maker posts its initial ladder around $v$. At step~1 the leveraged trader builds its position; at step~4 the chosen intervention takes effect; at step~7 the pre-shock reference $m_{t_0-1}$ is recorded; and at step $t_0=8$ the shock account sells 160--240 lots at market. Within each step, operations execute in a fixed order (Figure~\ref{fig:episode}b): the maker re-quotes; last step's shadow trades are settled and strategy shares updated; intervention orders are placed; the shock and its persistent impact are applied; forced liquidations execute; noise, fundamental, and trend orders arrive; the adaptive participant trades; and the end-of-step reference price and objective are recorded.

\begin{figure}[t]
\centering
\begin{tikzpicture}[x=1cm,y=1cm]
\node[anchor=west,font=\small\bfseries,text=ink] at (0,4.05) {(a) Episode timeline and scored drawdown (conceptual path)};
\draw[ink!60] (.8,.3) -- (16.3,.3);
\draw[ink!60] (.8,.3) -- (.8,2.75);
\node[lbl,rotate=90,anchor=south] at (.75,1.7) {reference price $m_t$};
\foreach \s/\x in {0/.8,1/1.3,4/2.8,7/4.3,8/4.8,32/7.91,64/12.05,96/16.2}{
  \draw[ink!60] (\x,.3) -- (\x,.22);
  \node[font=\scriptsize,text=ink,anchor=north] at (\x,.2) {\s};
}
\node[lbl,anchor=north] at (10.5,-.08) {step (pre-shock steps expanded)};
\fill[ask!22] (4.8,2.3) .. controls (5.1,2.25) and (5.6,.85) .. (6.2,.8)
  .. controls (7.4,.75) and (11.5,1.6) .. (16.2,1.9) -- (16.2,2.3) -- cycle;
\draw[ink!50,dashed] (4.3,2.3) -- (16.25,2.3);
\node[lbl,anchor=south east] at (16.25,2.32) {pre-shock reference $m_{t_0-1}$};
\draw[ink,line width=1pt] (.8,2.32) .. controls (1.6,2.27) and (2.4,2.35) .. (3.2,2.3)
  .. controls (3.8,2.26) and (4.4,2.33) .. (4.8,2.3)
  .. controls (5.1,2.25) and (5.6,.85) .. (6.2,.8)
  .. controls (7.4,.75) and (11.5,1.6) .. (16.2,1.9);
\node[font=\scriptsize,text=ask!85!black,align=center] at (9.6,1.75) {scored drawdown $\sum_{t\ge t_0}\max\{0,\,m_{t_0-1}-m_t\}$};
\foreach \x/\y/\txt/\c in {1.3/3.35/exposure build/ink!60, 2.8/2.95/intervention/cIv, 4.3/3.35/reference/ink!60, 4.8/2.95/shock/ask!85!black}{
  \draw[\c,densely dotted] (\x,.3) -- (\x,\y-.1);
  \node[font=\scriptsize,text=\c,anchor=south] at (\x,\y-.12) {\txt};
}
\node[anchor=west,font=\small\bfseries,text=ink] at (0,-.85) {(b) Order of operations within each step};
\tikzset{stp/.style={draw=ink!55,fill=white,rounded corners=1.5pt,align=center,font=\scriptsize,minimum height=.95cm,text width=2.62cm,inner sep=2pt}}
\foreach \i/\x/\y/\txt/\c in {
  1/1.45/-1.85/{Maker re-quotes\\(stress-scaled depth)}/bid,
  2/4.75/-1.85/{Settle shadow P\&L,\\update shares}/cAd,
  3/8.05/-1.85/{Intervention orders\\(deleveraging, backstop bid)}/cIv,
  4/11.35/-1.85/{Shock order and\\persistent impact}/ask,
  5/14.65/-1.85/{Margin check,\\forced liquidation}/ask,
  6/1.45/-3.25/{Noise orders\\($8\times4$ lots)}/ink!45,
  7/4.75/-3.25/{Fundamental and\\trend traders}/ink!45,
  8/8.05/-3.25/{Adaptive trader: fork\\shadows, draw, trade}/cAd,
  9/11.35/-3.25/{Record $m_t$ and\\objective}/ink!45}{
  \node[stp] (s\i) at (\x,\y) {\txt};
  \fill[\c] ($(s\i.north west)+(.03,-.03)$) rectangle ($(s\i.south west)+(.13,.03)$);
  \node[circle,fill=ink,text=white,font=\tiny\bfseries,inner sep=1pt,minimum size=9pt] at (s\i.north west) {\i};
}
\foreach \a/\b in {1/2,2/3,3/4,4/5,6/7,7/8,8/9}{\draw[->,ink!50] (s\a) -- (s\b);}
\draw[->,ink!50,rounded corners=3pt] (s5.south) -- ++(0,-.3) -| ($(s6.north)+(0,0)$);
\foreach \c/\t/\yy in {bid/liquidity/-2.75,cAd/adaptation/-3.05,cIv/intervention/-3.35,ask/shock and stress/-3.65,ink!45/order flow and bookkeeping/-3.95}{
  \fill[\c] (13.2,\yy-.08) rectangle (13.32,\yy+.08);
  \node[lbl,anchor=west] at (13.38,\yy) {\t};
}
\end{tikzpicture}
\caption{\textbf{Episode anatomy.} (a) Fixed schedule of a 96-step episode. The utility's drawdown term is the shaded area between the pre-shock reference and the post-shock reference path (the path drawn is illustrative, not a sampled trajectory). (b) Deterministic order of operations within a step. Shadow-trade settlement (2) and forking (8) run only when the strategy mix evolves; margin checks (5), fundamental and trend traders (7), and the adaptive participant (8) are active from the \emph{leverage} stage onward (Figure~\ref{fig:mechanisms}a).}
\label{fig:episode}
\end{figure}

\subsection{Amplification mechanisms}
\label{sec:amplification}
The shock is a burst of order flow, not a change in fundamentals. Three mechanisms convert it into a crash whose depth and duration depend on the world (Figure~\ref{fig:mechanisms}b).

\textbf{Persistent impact.} The maker's quote center shifts by $40\,x/D_0$ ticks, where $x$ is the executed shock volume and $D_0=10\bar Q$ is the depth of the ten best bids in the initial (step-0) ladder; integer remainders are carried forward. The shock therefore moves the price permanently rather than only transiently.

\textbf{Liquidity withdrawal.} After the shock, the maker measures stress $\sigma_t=\max\{0,\,m_{t_0-1}-m_{t-1}\}+|r_{t-1}|$ in half-ticks and thins its ladder to
\begin{equation}
Q_t=\bar Q\cdot\max\bigl\{0.25,\;1-\beta\,\sigma_t/\kappa\bigr\},
\label{eq:withdrawal}
\end{equation}
where $\beta\in[0,1]$ is the world's withdrawal strength and $\kappa\in\{24,\dots,56\}$ its stress scale. Withdrawal reduces depth but never widens the spread or empties the book.

\textbf{Margin-driven liquidation.} At the start of step $t$, the leveraged trader's equity is $E_t=c_t+q_t p_t$, where $c_t$ and $q_t$ are its cash and inventory at the close of step $t-1$ and $p_t=m_{t-1}/2$ is the previous closing midpoint in ticks; its maintenance requirement is $M_t=\mu\,|q_t|\,p_t$ with $\mu\in[3\%,7\%]$. From step $t_0+1$ onward, whenever $E_t<M_t$ it must sell
\begin{equation}
\ell_t=\bigl\lceil \alpha\,|q_t|\bigr\rceil
\label{eq:margin}
\end{equation}
lots at market, where $\alpha\in[0,1]$ is the world's liquidation strength. Because forced sales lower the price and therefore equity, the rule can repeat over consecutive steps and form a liquidation spiral.

The strengths $\alpha$ and $\beta$ are both held at zero until the shock order has filled, and they define the four world strata used in sampling: neither mechanism, liquidation only, withdrawal only, or both (Section~\ref{sec:population}).

\subsection{Reflexive strategy ecology}
\label{sec:ecology}
The adaptive participant holds shares $(s_{\mathrm F},s_{\mathrm{Tr}})$ over two candidate strategies, initialized at $(\tfrac12,\tfrac12)$. Both candidates are inactive until the shock order has filled. The \textsc{Fundamental} candidate buys $\min\{80,\,4(v-\text{ask}-\text{fee})\}$ lots when the ask is below value net of fees, sells symmetrically when the bid is above it, and otherwise does nothing. The \textsc{Trend} candidate is a fixed, signal-independent order pair that sells 160 lots and then buys back 80, a net sale of 80 lots. Each step after the shock, the participant draws one candidate with probabilities $(s_{\mathrm F},s_{\mathrm{Tr}})$ and submits it at market.

When the strategy mix evolves, the simulator first executes \emph{shadow trades}: at the point just before the adaptive participant's real order, each candidate is executed on its own copy of the exchange. At the start of the next step, after the maker re-quotes, each shadow position is marked to the refreshed midpoint, giving one-step counterfactual profits $\pi_{\mathrm F},\pi_{\mathrm{Tr}}$ (net of fees, in half-tick cash units). Shares are then updated by a multiplicative replicator rule,
\begin{equation}
s_i'\;\propto\;s_i\bigl(1+\eta\,(\pi_i-\textstyle\min_k\pi_k)\bigr),\qquad \eta=10^{-3},
\label{eq:replicator}
\end{equation}
with exact integer normalization in parts per million. An intervention therefore affects not only the current order book but also the strategy mix that generates future order flow. When the mix is fixed, the adaptive participant still trades, but shadow trades and Eq.~\eqref{eq:replicator} are disabled and the shares stay at $(\tfrac12,\tfrac12)$.

\subsection{Switchable mechanism stages}
Mechanisms are enabled cumulatively in five stages (Figure~\ref{fig:mechanisms}a): a \emph{mechanical} stage with only maker, noise, leveraged exposure, shock, and interventions; \emph{+impact}, which adds persistent impact; \emph{+withdrawal}, which adds liquidity withdrawal; \emph{+leverage}, which adds margin-driven liquidation, the fundamental and trend traders, and the adaptive participant with a fixed mix; and \emph{+evolution}, which adds shadow trades and replicator updates. Our experiments use two laboratories whose dynamics differ in exactly one mechanism. The \textbf{fixed-mix laboratory} runs the \emph{+leverage} stage. The \textbf{adaptive laboratory} runs the full \emph{+evolution} stage, which is also the configuration of every target. The first three stages are implemented but not used in the experiments reported here.

\begin{figure}[t]
\centering
\begin{tikzpicture}[x=1cm,y=1cm,
  nd/.style={draw=ink!55,fill=white,rounded corners=3pt,align=center,font=\footnotesize,
             inner sep=3pt,minimum height=.8cm,text width=#1,line width=.6pt},
  nd/.default=1.75cm,
  loopL/.style={->,line width=.9pt,ask!80!black},
  loopW/.style={->,line width=.9pt,bid!80!black},
  evo/.style={->,line width=.8pt,cAd},
  plabel/.style={anchor=west,font=\small\bfseries,text=ink},
]
\node[plabel] at (0,5.55) {(a) Mechanisms enabled in each stage};
\fill[cFx!13,rounded corners=4pt] (12.25,-.15) rectangle (14.15,5.05);
\fill[cAd!13,rounded corners=4pt] (14.45,-.15) rectangle (16.35,5.05);
\foreach \i in {1,3,5}{\fill[ink!4] (0,4.55-\i*.66+.33) rectangle (16.35,4.55-\i*.66-.33);}
\foreach \j/\name in {1/Mechanical,2/+impact,3/+withdrawal,4/+leverage,5/+evolution}{
  \node[font=\footnotesize\bfseries,text=ink] at (4.4+\j*2.2,4.72) {\name};}
\foreach \i/\name in {1/Persistent impact of the shock,2/Liquidity withdrawal (Eq.~\ref*{eq:withdrawal}),3/Margin-driven liquidation (Eq.~\ref*{eq:margin}),4/Fundamental and trend traders,5/Adaptive participant (fixed mix),6/Shadow trades + replicator (Eq.~\ref*{eq:replicator})}{
  \node[anchor=west,font=\footnotesize,text=ink] at (.15,4.55-\i*.66) {\name};}
\foreach \i/\first in {1/2,2/3,3/4,4/4,5/4,6/5}{
  \foreach \j in {1,...,5}{
    \ifnum\j<\first
      \draw[ink!28,line width=.6pt] (4.4+\j*2.2,4.55-\i*.66) circle[radius=3.6pt];
    \else
      \ifnum\j=5 \fill[cAd] (4.4+\j*2.2,4.55-\i*.66) circle[radius=4pt];
      \else\ifnum\j=4 \fill[cFx] (4.4+\j*2.2,4.55-\i*.66) circle[radius=4pt];
      \else \fill[ink!80] (4.4+\j*2.2,4.55-\i*.66) circle[radius=4pt];\fi\fi
    \fi}}
\draw[cAd,line width=.7pt,rounded corners=3pt,dashed] (12.3,4.55-6*.66-.29) rectangle (16.3,4.55-6*.66+.29);
\node[font=\scriptsize\bfseries,text=cFx!85!black,align=center] at (13.2,-.55) {fixed-mix\\laboratory};
\node[font=\scriptsize\bfseries,text=cAd!85!black,align=center] at (15.4,-.55) {adaptive lab.\\and target};
\draw[ink!35,line width=.6pt] (6.0,-.32) -- (6.0,-.42) -- (11.6,-.42) -- (11.6,-.32);
\node[font=\scriptsize,text=muted] at (8.8,-.68) {implemented; not used in these experiments};
\node[anchor=west,font=\scriptsize,text=muted] at (.15,-.5) {All stages: maker, noise flow, leveraged exposure,};
\node[anchor=west,font=\scriptsize,text=muted] at (.15,-.82) {shock, and both interventions.};
\node[plabel] at (0,-1.55) {(b) Amplification loops and intervention entry points};
\begin{scope}[shift={(4.05,-4.85)}]
\node[nd=1.6cm,draw=ink!80,line width=.8pt,fill=ink!5] (pd) at (0,0) {\textbf{Price}\\\textbf{decline}};
\node[nd=1.6cm] (sh) at (0,1.75) {Shock sale\\(step 8)};
\node[nd] (md) at (2.55,.85) {Margin\\deficit};
\node[nd] (fs) at (2.55,-.85) {Forced\\selling};
\node[nd] (ms) at (-2.55,.85) {Maker\\stress};
\node[nd] (tl) at (-2.55,-.85) {Thinner\\ladder};
\draw[->,line width=.9pt,ink!75] (sh) -- (pd);
\draw[loopL] (pd.north east) to[out=30,in=180] (md.west);
\draw[loopL] (md) -- node[right,font=\footnotesize,text=ask!80!black]{$\alpha$} (fs);
\draw[loopL] (fs.west) to[out=180,in=-30] (pd.south east);
\draw[loopW] (pd.north west) to[out=150,in=0] (ms.east);
\draw[loopW] (ms) -- node[left,font=\footnotesize,text=bid!80!black]{$\beta$} (tl);
\draw[loopW] (tl.east) to[out=0,in=210] (pd.south west);
\node[font=\scriptsize\bfseries,text=ask!80!black] at (2.55,-1.62) {liquidation loop};
\node[font=\scriptsize\bfseries,text=bid!80!black] at (-2.55,1.62) {withdrawal loop};
\node[nd=2.9cm,draw=cIv,fill=cIv!8,font=\scriptsize] (cap) at (2.55,2.25) {\texttt{LEVERAGE\_CAP}\\sell half, step 4};
\node[nd=2.9cm,draw=cIv,fill=cIv!8,font=\scriptsize] (bs) at (-2.55,-2.25) {\texttt{LIQUIDITY\_BACKSTOP}\\40-lot bid each step};
\draw[-{Bar[width=7pt]},line width=1.1pt,cIv] (cap) -- (md);
\draw[-{Bar[width=7pt]},line width=1.1pt,cIv] (bs) -- (tl);
\end{scope}
\node[plabel] at (9.35,-1.55) {(c) Evolving strategy mix (adaptive only)};
\begin{scope}[shift={(10.3,-4.85)}]
\node[nd=1.85cm,draw=ink!80,fill=ink!5] (ex) at (0,1.35) {Exchange\\at step $t$};
\node[nd=1.75cm,draw=cAd,fill=cAd!7] (sf) at (2.75,1.9) {shadow \textsc{Fund.}};
\node[nd=1.75cm,draw=cAd,fill=cAd!7] (st) at (2.75,.8) {shadow \textsc{Trend}};
\node[nd=1.75cm,draw=cAd,fill=cAd!7] (mk) at (5.0,1.35) {mark at mid\\of step $t{+}1$};
\node[nd=2.75cm,draw=cAd,fill=cAd!7] (up) at (3.3,-1.05) {replicator update\\$s_i'\propto s_i(1{+}\eta\,\Delta\pi_i)$};
\node[nd=1.85cm] (dr) at (0,-1.05) {draw strategy,\\real order};
\draw[evo] (ex.east) to[out=0,in=180] (sf.west);
\draw[evo] (ex.east) to[out=0,in=180] (st.west);
\draw[evo] (sf.east) to[out=0,in=180] (mk.west);
\draw[evo] (st.east) to[out=0,in=180] (mk.west);
\draw[evo] (mk.south) |- node[pos=.25,right,font=\scriptsize,text=cAd!85!black]{$\pi_{\mathrm F},\pi_{\mathrm{Tr}}$} (up.east);
\draw[evo] (up.west) -- node[above,font=\scriptsize,text=muted]{$s$} (dr.east);
\draw[->,line width=.8pt,ink!75] (dr.north) -- node[left,font=\scriptsize,text=muted,align=right]{order\\flow} (ex.south);
\node[font=\scriptsize,text=muted,align=center] at (2.4,-2.2) {an intervention changes $\pi_i$, hence the future mix};
\end{scope}
\end{tikzpicture}
\caption{\textbf{Switchable mechanisms.} (a) Mechanisms are enabled cumulatively; filled circles mark the active ones. The two laboratories differ only in the outlined last row. (b) The shock's price decline feeds a liquidation loop (red; strength $\alpha$) and a withdrawal loop (teal; strength $\beta$). The deleveraging intervention shrinks the leveraged inventory before the shock and so caps the margin deficit; the backstop's resting bid offsets the thinning of the ladder. (c) With an evolving mix, each step forks both candidate strategies onto exchange copies, marks them one step later, and re-weights the mix ($\Delta\pi_i=\pi_i-\min_k\pi_k$).}
\label{fig:mechanisms}
\end{figure}

\subsection{Determinism, replay, and seed hygiene}
\label{sec:replay}
\textbf{Counter-based randomness.} Random numbers are not drawn from a stateful generator. Each draw is the top 53 bits of an HMAC-SHA256 of its address---namespace, step, entity, and draw index---under a key derived from the seed. Draws are therefore independent of call order and of the market state. Every step consumes exactly ten draws (one reserved for fundamental innovations, which are disabled in this construction; eight for noise traders; and one for the adaptive participant), and each draw is recorded and recomputed during verification. All arms that share a seed, such as the two interventions in the same repeat, thus face identical random inputs: their contrasts are evaluated under common random numbers.

\textbf{Event sourcing and replay.} The exchange appends every order acceptance, rejection, fill, rest, cancellation, and cancel rejection to an event log with contiguous identifiers and non-decreasing steps, and maintains a running SHA-256 over its canonical serialization. A replayer rebuilds the full state from events alone and re-validates risk admission, price--time priority, resting prices, fill quantities, counterparties, and fees.

\textbf{Seed hygiene.} Seeds are derived by hashing a domain tag, master entropy, the world constructor's hash, the cohort, and the repeat index. Only a commitment to each seed is stored; seeds that collide with any previously used commitment are re-derived. In the prospective study this yields 384 unique seeds (24 worlds $\times$ 2 cohorts $\times$ 8 repeats), and each seed is simulated for ten run roles (initial observation, control, three paired experiments, and two interventions) in both laboratory configurations: 7,680 logical runs, or 4,608 after deduplicating identical executions.

\section{The decision task}
\label{sec:task}

\subsection{World population}
\label{sec:population}
A world is a parameterization of the crash construction. Each world draws a liquidation strength $\alpha$ and a withdrawal strength $\beta$ uniformly from $[0.05,1]$ and then sets the inactive one to zero according to its stratum: neither mechanism, liquidation only, withdrawal only, or both. It also draws the shock size (160--240 lots), the leveraged exposure (160--240 lots), the leveraged trader's cash (84k--124k, raised if necessary to satisfy the $20\times$ leverage limit), the maintenance margin $\mu$, the maker depth $\bar Q$, and the stress scale $\kappa$ (Appendix~\ref{app:params}). Candidate worlds are rejected only for schema violations or duplicate constructors, never on the basis of outcomes.

\subsection{Interventions}
\label{sec:interventions}
The agent chooses one of two interventions, both taking effect at step~4, four steps before the shock.
\begin{itemize}[leftmargin=1.4em,itemsep=1pt,topsep=2pt]
\item \act{LEVERAGE\_CAP} (deleveraging): a one-time deleveraging sale in which the leveraged trader sells $\lceil |q|/2\rceil$ lots at market, halving its position; the intervention account also buys two lots at market. Despite its identifier, it imposes no lasting leverage limit.
\item \act{LIQUIDITY\_BACKSTOP}: the intervention account buys one lot at market and then, at every step from step~4 onward, posts a single resting 40-lot bid at the best bid, cancelling the previous step's unfilled bid.
\end{itemize}
An intervention is \emph{feasible} in a world if, in all eight target repeats, every scheduled marketable order (exposure build, intervention trades, shock, and forced liquidations) executes in full, every backstop bid rests, and no order is rejected by risk admission. Regret is defined only for feasible chosen actions.

\subsection{Utility and regret}
For one native repeat, the utility of intervention $a$ is
\begin{equation}
U(a)=-\sum_{t=t_0}^{T}\max\{0,\;m_{t_0-1}-m_t(a)\}\;-\;\frac{C_{\mathrm I}(a)+C_{\mathrm L}(a)}{\lambda},
\label{eq:utility}
\end{equation}
with $t_0=8$, $T=96$, and $m_t$ the reference price in half-ticks. The first term is the area by which the reference price falls below its pre-shock level (Figure~\ref{fig:episode}a). The second term charges implementation costs inside the market: $C_{\mathrm I}$ is the fees paid by the intervention account (including backstop fills), $C_{\mathrm L}$ the fees paid by the leveraged trader on the deleveraging sale, and $\lambda=100$ a fixed cost scale. These are market costs of the intervention, not costs of querying the agent. Let $\mathcal A_{w,1-j}$ be the feasible interventions in target cohort $1-j$ and $\bar U_{w,1-j}(a)$ the mean of~\eqref{eq:utility} over its eight repeats. The regret of a decision made with laboratory cohort $j$ is
\begin{equation}
\R_{w,j}(a)=\max_{a'\in\mathcal A_{w,1-j}}\bar U_{w,1-j}(a')-\bar U_{w,1-j}(a).
\label{eq:regret}
\end{equation}
With two feasible interventions, regret is zero for an optimal choice and equals the target utility gap otherwise. Lower mean regret therefore means avoiding \emph{costly} mistakes, not merely making more correct choices. Because every feasible intervention is executed on the same target repeats, Eq.~\eqref{eq:regret} is exact counterfactual ground truth for those eight realized repeats; it is a finite-sample estimate of the world's expected regret. Its units are those of the simulator objective; a monetary interpretation would require external calibration.

\subsection{Experiments and observations}
\textbf{Experiment menu.} Each experiment is a paired comparison on the eight laboratory repeats, using the same seed for both arms. \act{LEVERAGE} compares the deleveraging arm with an otherwise identical arm whose deleveraging trade is removed; \act{LIQUIDITY} does the same for the backstop bids; and \act{NULL} compares the control arm with itself. Each experiment therefore isolates the effect of the intervention's market action, but the two experiments use different controls, so their effects are not directly on a common baseline.

\textbf{Summary packets.} An experiment returns an aggregate over the eight repeats of the treated arm's public statistics---return, spread, volume, depth, reference drawdown, order flow, price impact, strategy and adaptation statistics, quote-availability statistics, and counts of forced liquidations and maker order cancellations---together with the paired utility difference (mean, 5th/50th/95th percentiles, and a per-step effect size). Packets never contain target-cohort outcomes.

\textbf{Initial observation.} Every condition begins with a public message of the same form: an eight-repeat summary of a \emph{shock-free} control run on the laboratory seeds, simulated in the condition's laboratory configuration; the public objective; the names of the two candidate amplification mechanisms and a graph of their possible causal paths to utility; the transfer context (laboratory configuration, adaptive target, disjoint randomness); and the experiment budget and menu. Across conditions, only the stated laboratory configuration and the budget and menu (none under \Obs) differ. Because the initial run contains no shock, an agent without experiments must infer the crash dynamics from this structural description alone.

\section{Experimental protocol}
\label{sec:protocol}

\subsection{Information conditions}
The agent interacts through two strict JSON tools, \texttt{run\_experiment} and \texttt{submit\_decision}. All conditions are scored on adaptive targets:
\begin{itemize}[leftmargin=1.4em,itemsep=1pt,topsep=2pt]
\item \Obs: the initial observation only; \texttt{run\_experiment} is not offered.
\item \Ad: up to eight experiment calls in the adaptive laboratory.
\item \Fx: up to eight experiment calls in the fixed-mix laboratory.
\item \Pl: identical tooling, but every call returns one fixed packet---the \act{LEVERAGE} packet of a single previously examined world---with all paired-difference fields set to zero (prospective study only).
\end{itemize}
Each experiment's packet is prepared once; repeating a call returns the same packet and consumes budget, so eight calls do not imply eight independent experiments. The \Ad{}--\Fx{} contrast isolates the evolving strategy mix (Figure~\ref{fig:mechanisms}a) together with the stated laboratory configuration, rather than comparing a generic ``unrealistic'' market with a ``realistic'' one. Because the placebo packet contains real crash statistics from another world, \Pl{} separates world-specific evidence from tool interaction combined with generic crash information.

\subsection{Agents and interaction}
The primary agent is ChatGPT~5.6.4, evaluated in its three served variants Sol, Terra, and Luna (API model identifiers \texttt{gpt-5.6-sol}, \texttt{gpt-5.6-terra}, and \texttt{gpt-5.6-luna}). Each variant is analyzed as a separate deployment. GPT-5.5 is a supplementary deployment evaluated in the exploratory study only (Appendix~\ref{app:details}). Requests use temperature zero, one completion, a 4,096-token completion limit, and no parallel tool calls. An episode allows at most 32 turns with one tool call each. Once the experiment budget is spent, or during the last two of the 32 turns, only \texttt{submit\_decision} is offered and the request forces a tool call. The first legal submission is final. Malformed or illegal calls receive a deterministic error message and consume a turn; a third error is terminal. The agent's history contains only the public message, its own messages and tool calls, and tool results: target outcomes, feasibility, optimal actions, and correctness are never revealed, and completed nonoptimal decisions are never retried.

\subsection{Studies and statistical analysis}
\textbf{Prospective confirmation.} After the primary protocol, population, and analysis were fixed in an internal specification (not a public registry), we sampled 24 new worlds, excluding all previously examined constructors and seed commitments, with six worlds in each stratum. Within each world and cohort, the order of conditions was rotated. The design crosses 24 worlds, two transfer directions, four conditions, and three deployments, for 576 planned decisions.

\textbf{Exploratory study.} An earlier study used 24 previously examined worlds, two directions, and three conditions (\Obs, \Ad, \Fx) for four deployments, again 576 decisions, with the same tool interface. Its intervals were specified after observing outcomes and are descriptive.

\textbf{Primary contrast.} For each deployment the primary contrast is the paired mean
$\Delta=\operatorname{mean}_{w,j}\bigl[\R^{\AdS}_{w,j}-\R^{\ObsS}_{w,j}\bigr]$ over the 48 world--direction units. Uncertainty is estimated with 10,000 bootstrap resamples of whole worlds within strata~\citep{efron1979,cameron2008}, preserving both directions and all conditions of each resampled world. The prespecified criterion is met when $\Delta<0$, the nominal 95\% interval lies entirely below zero, and all 48 pairs are available. Deployments are analyzed separately; we make no pooled or multiplicity-adjusted claim.

\section{Results}
\label{sec:results}

\subsection{Prospective confirmation on new worlds}
\begin{figure}[t]
\centering
\begin{tikzpicture}
\begin{axis}[msb,
  title={(a) Mean target regret, Luna},
  name=ma,scale only axis,
  width=4.3cm,height=4.0cm,
  xmin=0,xmax=11,ymin=.45,ymax=4.55,
  ytick={1,2,3,4},
  yticklabels={\Fx,\Ad,\Pl,\Obs},
  y tick label style={font=\footnotesize},
  xtick={0,2,4,6,8,10},
  xlabel={Mean regret (lower is better)},
  axis x line*=bottom,axis y line*=left,
  y axis line style={draw=none},ytick style={draw=none},
  xmajorgrids,grid style={ink!9},
  clip=false,
]
\fill[cOb!88,rounded corners=1.5pt] (axis cs:0,3.73) rectangle (axis cs:8.878,4.27);
\node[anchor=west,font=\scriptsize,text=ink] at (axis cs:8.878,4) {\,\textbf{8.878}\;{\color{muted}(30/48)}};
\fill[cPl!88,rounded corners=1.5pt] (axis cs:0,2.73) rectangle (axis cs:6.509,3.27);
\node[anchor=west,font=\scriptsize,text=ink] at (axis cs:6.509,3) {\,\textbf{6.509}\;{\color{muted}(33/48)}};
\fill[cAd!88,rounded corners=1.5pt] (axis cs:0,1.73) rectangle (axis cs:3.785,2.27);
\node[anchor=west,font=\scriptsize,text=ink] at (axis cs:3.785,2) {\,\textbf{3.785}\;{\color{muted}(37/48)}};
\fill[cFx!88,rounded corners=1.5pt] (axis cs:0,0.73) rectangle (axis cs:2.66,1.27);
\node[anchor=west,font=\scriptsize,text=ink] at (axis cs:2.66,1) {\,\textbf{2.660}\;{\color{muted}(44/48)}};
\end{axis}
\begin{axis}[msb,
  title={(b) Paired differences},
  at={(ma.east)},anchor=west,xshift=3.0cm,
  scale only axis,width=4.3cm,height=4.0cm,
  xmin=-11.5,xmax=4.5,ymin=.45,ymax=5.55,
  ytick={1,2,3,4,5},
  yticklabels={\AdS$-$\FxS,\AdS$-$\PlS,\PlS$-$\ObsS,\FxS$-$\ObsS,\textbf{\AdS$-$\ObsS}},
  y tick label style={font=\footnotesize},
  xtick={-10,-5,0},
  xlabel={Regret difference ($<0$ favors first)},
  axis x line*=bottom,axis y line*=left,
  y axis line style={draw=none},ytick style={draw=none},
  xmajorgrids,grid style={ink!9},
  clip=false,
]
\fill[cAd!9,rounded corners=2pt] (axis cs:-11.5,4.55) rectangle (axis cs:4.5,5.45);
\draw[ink!60,dashed,line width=.6pt] (axis cs:0,.45) -- (axis cs:0,5.55);
\node[anchor=west,font=\scriptsize\bfseries,text=muted] at (axis cs:5.0,5.85) {mean};
\node[anchor=west,font=\scriptsize\bfseries,text=muted] at (axis cs:8.6,5.85) {95\% interval};
\draw[cAd,line width=1.4pt,line cap=round] (axis cs:-9.507,5) -- (axis cs:-1.028,5);
\fill[cAd,draw=white,line width=.9pt] (axis cs:-5.092,5) circle[radius=3.6pt];
\node[anchor=west,font=\scriptsize,text=ink] at (axis cs:5.0,5) {$\mathbf{-5.09}$};
\node[anchor=west,font=\scriptsize,text=ink] at (axis cs:8.6,5) {$[-9.51,\,-1.03]$};
\draw[cFx,line width=1.4pt,line cap=round] (axis cs:-10.645,4) -- (axis cs:-2.269,4);
\fill[cFx,draw=white,line width=.9pt] (axis cs:-6.218,4) circle[radius=3.6pt];
\node[anchor=west,font=\scriptsize,text=ink] at (axis cs:5.0,4) {$-6.22$};
\node[anchor=west,font=\scriptsize,text=ink] at (axis cs:8.6,4) {$[-10.65,\,-2.27]$};
\draw[cPl,line width=1.4pt,line cap=round] (axis cs:-6.145,3) -- (axis cs:1.36,3);
\fill[cPl,draw=white,line width=.9pt] (axis cs:-2.369,3) circle[radius=3.6pt];
\node[anchor=west,font=\scriptsize,text=ink] at (axis cs:5.0,3) {$-2.37$};
\node[anchor=west,font=\scriptsize,text=ink] at (axis cs:8.6,3) {$[-6.15,\,1.36]$};
\draw[ink!55,line width=1.4pt,line cap=round] (axis cs:-6.263,2) -- (axis cs:0.972,2);
\fill[ink!55,draw=white,line width=.9pt] (axis cs:-2.724,2) circle[radius=3.6pt];
\node[anchor=west,font=\scriptsize,text=ink] at (axis cs:5.0,2) {$-2.72$};
\node[anchor=west,font=\scriptsize,text=ink] at (axis cs:8.6,2) {$[-6.26,\,0.97]$};
\draw[ink!55,line width=1.4pt,line cap=round] (axis cs:-1.983,1) -- (axis cs:3.756,1);
\fill[ink!55,draw=white,line width=.9pt] (axis cs:1.126,1) circle[radius=3.6pt];
\node[anchor=west,font=\scriptsize,text=ink] at (axis cs:5.0,1) {$1.13$};
\node[anchor=west,font=\scriptsize,text=ink] at (axis cs:8.6,1) {$[-1.98,\,3.76]$};
\end{axis}
\end{tikzpicture}
\caption{\textbf{Laboratory access lowers Luna's regret relative to observation alone} (prospective study, 24 worlds). (a) Mean target regret per condition; parentheses give optimal-action counts. (b) Paired mean differences with nominal 95\% stratified world-bootstrap intervals (\ObsS{} = \Obs, \AdS{} = \Ad, \FxS{} = \Fx, \PlS{} = \Pl). The shaded row is the prespecified primary contrast. Among the secondary contrasts, the fixed-mix laboratory also improves on \Obs, whereas the contrasts against the placebo and between the two laboratories are inconclusive.}
\label{fig:effects}
\end{figure}
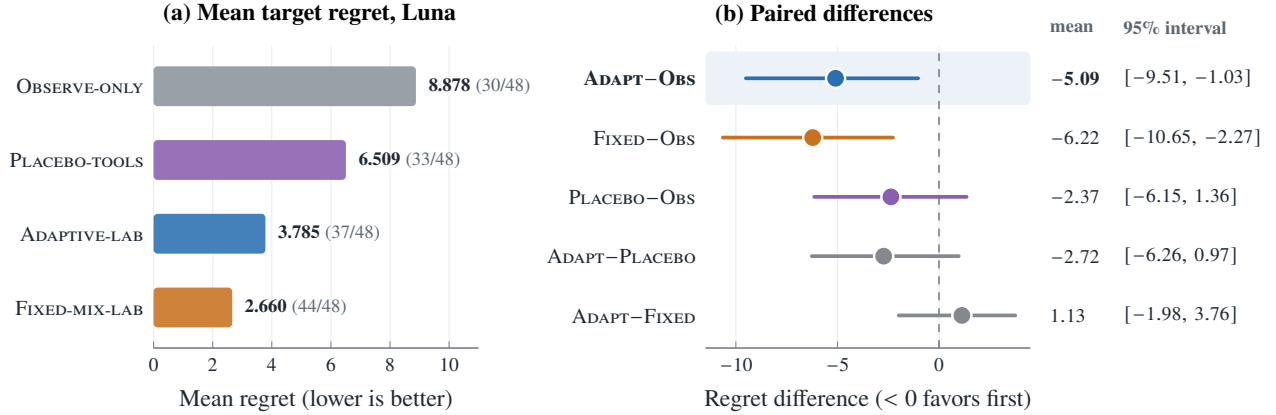

Of the 576 planned prospective decisions, 362 were scored, 123 ended without an action (empty outputs), and 91 Sol decisions were blocked by an infrastructure failure. Luna completed all 192 of its decisions. Its mean regret falls from 8.878 under \Obs{} to 3.785 under \Ad: $\Delta=-5.092$, with nominal 95\% interval $[-9.507,-1.028]$ (Table~\ref{tab:luna}, Figure~\ref{fig:effects}). This is a 57.4\% reduction in mean regret and meets the prespecified primary criterion.

Sol and Terra produced 88 and 82 scored decisions of 192, leaving only 20 and 16 of 48 complete primary pairs, respectively. Their planned primary comparisons therefore remain unresolved; we do not substitute complete-case estimates for them.

\begin{table}[t]
\centering
\small
\begin{tabular}{lcccc}
\toprule
Condition & Mean regret $\downarrow$ & Optimal actions & Scored / planned & Difference from \Obs\\
\midrule
\Obs & 8.878 & 30/48 & 48/48 & ---\\
\Pl & 6.509 & 33/48 & 48/48 & $-2.369$\\
\Ad{} (primary) & \textbf{3.785} & 37/48 & 48/48 & $\mathbf{-5.092}$\\
\Fx & 2.660 & 44/48 & 48/48 & $-6.218$\\
\bottomrule
\end{tabular}
\caption{\textbf{Prospective results for Luna} on 24 newly sampled worlds (48 world--direction units). Optimal-action counts include any feasible utility maximizer. The last column is the paired mean difference from \Obs; only \Ad{} versus \Obs{} is the prespecified primary contrast, and the others are secondary (intervals in Figure~\ref{fig:effects}b).}
\label{tab:luna}
\end{table}

\subsection{What the controls establish}
\textbf{Placebo.} For Luna, the secondary \AdS$-$\PlS{} difference is $-2.724$ with interval $[-6.263,0.972]$, and \PlS$-$\ObsS{} is $-2.369$ with interval $[-6.145,1.360]$. The prespecified secondary \FxS$-$\ObsS{} contrast is $-6.218$ with interval $[-10.645,-2.269]$, so the fixed-mix laboratory also improves on observation alone. Its post hoc contrast with the placebo, $\FxS-\PlS=-3.849$, has interval $[-8.395,1.158]$ under the same bootstrap draws. These controls leave uncertain how much of the benefit comes from world-specific experimental evidence rather than from tool interaction and generic crash information. The primary result concerns the complete laboratory-access condition relative to \Obs.

\textbf{Laboratory design.} An evolving strategy mix in the laboratory shows no clear advantage. Luna's prospective \AdS$-$\FxS{} difference is $1.126$ with interval $[-1.983,3.756]$, and all four exploratory \AdS$-$\FxS{} intervals also include zero (Table~\ref{tab:exploratory}). The interval admits both a benefit and a cost of the evolving mix; it does not establish that the two laboratories are equivalent.

\textbf{Binary outcomes.} Regret weights each mistaken choice by its target utility gap. Luna's optimal-action count also rises from 30 to 37 of 48, but the prespecified binary companion tests do not meet their directional criterion (Appendix~\ref{app:details}). The primary evidence therefore concerns avoided loss rather than accuracy.

\textbf{Randomization check.} Percentile intervals from 24 resampled clusters can be anti-conservative, so we add a post hoc exact sign-flip test. It enumerates all $2^{24}$ sign assignments of the 24 world-level mean differences. For \AdS$-$\ObsS, 13 worlds favor \Ad, 7 favor \Obs, and 4 tie; the test gives $p=0.037$ in the prespecified direction and $p=0.075$ two-sided. For \FxS$-$\ObsS{} the corresponding values are $p=0.009$ and $p=0.018$. The direction of the primary effect is therefore robust to the inference method, but its two-sided strength is borderline at 24 worlds.

\textbf{How the agent used the laboratory.} Luna made 4.71 experiment calls on average with \Ad{} (range 4--7), 4.52 with \Fx{} (2--6), and 4.25 with \Pl{} (3--7), out of eight. Because the menu has three experiments and repeated calls return the same packet, the budget always sufficed to observe the complete menu, and no Luna episode exhausted it. The improvement thus does not come from more calls: the placebo episodes were similar in length but returned no world-specific evidence. Table~\ref{tab:census} reports calls and outcomes for every deployment and condition.

\subsection{Exploratory evidence across deployments}
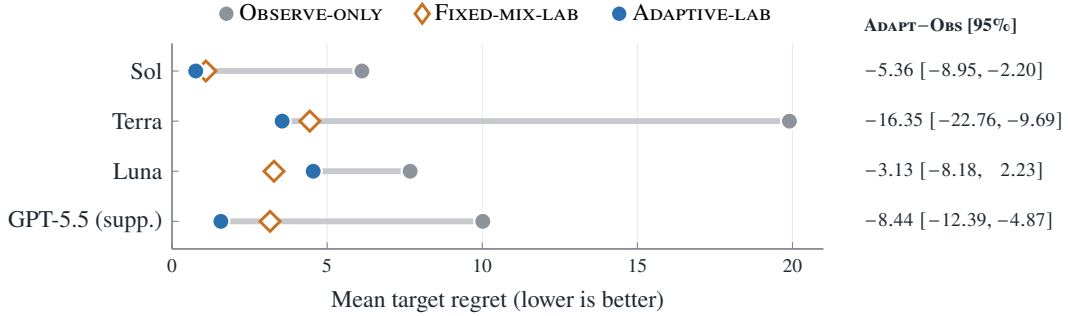
\begin{figure}[t]
\centering
\begin{tikzpicture}
\begin{axis}[msb,
  width=10.2cm,height=4.3cm,
  xmin=0,xmax=21,
  ymin=.45,ymax=4.55,
  ytick={1,2,3,4},
  yticklabels={GPT-5.5 (supp.),Luna,Terra,Sol},
  y tick label style={font=\small},
  xtick={0,5,10,15,20},
  xlabel={Mean target regret (lower is better)},
  axis x line*=bottom,axis y line*=left,
  xmajorgrids,grid style={ink!10},
  clip=false,
  legend style={at={(.5,1.04)},anchor=south,draw=none,font=\small,legend columns=3,
                /tikz/every even column/.append style={column sep=10pt}},
  legend cell align=left,
]
\addlegendimage{only marks,mark=*,mark size=3pt,mark options={fill=cOb,draw=white}}
\addlegendentry{\Obs}
\addlegendimage{only marks,mark=diamond*,mark size=3.6pt,mark options={fill=white,draw=cFx,line width=1.1pt}}
\addlegendentry{\Fx}
\addlegendimage{only marks,mark=*,mark size=3pt,mark options={fill=cAd,draw=white}}
\addlegendentry{\Ad}
\draw[ink!25,line width=2.2pt] (axis cs:0.764,4) -- (axis cs:6.121,4);
\node[diamond,fill=white,draw=cFx,line width=1.1pt,inner sep=0pt,minimum size=8pt] at (axis cs:1.091,4) {};
\fill[cOb,draw=white,line width=.8pt] (axis cs:6.121,4) circle[radius=3.2pt];
\fill[cAd,draw=white,line width=.8pt] (axis cs:0.764,4) circle[radius=3.2pt];
\node[anchor=west,font=\scriptsize,text=ink] at (axis cs:22,4) {$-5.36\;[-8.95,-2.20]$};
\draw[ink!25,line width=2.2pt] (axis cs:3.55,3) -- (axis cs:19.902,3);
\node[diamond,fill=white,draw=cFx,line width=1.1pt,inner sep=0pt,minimum size=8pt] at (axis cs:4.441,3) {};
\fill[cOb,draw=white,line width=.8pt] (axis cs:19.902,3) circle[radius=3.2pt];
\fill[cAd,draw=white,line width=.8pt] (axis cs:3.55,3) circle[radius=3.2pt];
\node[anchor=west,font=\scriptsize,text=ink] at (axis cs:22,3) {$-16.35\;[-22.76,-9.69]$};
\draw[ink!25,line width=2.2pt] (axis cs:4.551,2) -- (axis cs:7.676,2);
\node[diamond,fill=white,draw=cFx,line width=1.1pt,inner sep=0pt,minimum size=8pt] at (axis cs:3.287,2) {};
\fill[cOb,draw=white,line width=.8pt] (axis cs:7.676,2) circle[radius=3.2pt];
\fill[cAd,draw=white,line width=.8pt] (axis cs:4.551,2) circle[radius=3.2pt];
\node[anchor=west,font=\scriptsize,text=ink] at (axis cs:22,2) {$-3.13\;[-8.18,\phantom{-}2.23]$};
\draw[ink!25,line width=2.2pt] (axis cs:1.575,1) -- (axis cs:10.017,1);
\node[diamond,fill=white,draw=cFx,line width=1.1pt,inner sep=0pt,minimum size=8pt] at (axis cs:3.16,1) {};
\fill[cOb,draw=white,line width=.8pt] (axis cs:10.017,1) circle[radius=3.2pt];
\fill[cAd,draw=white,line width=.8pt] (axis cs:1.575,1) circle[radius=3.2pt];
\node[anchor=west,font=\scriptsize,text=ink] at (axis cs:22,1) {$-8.44\;[-12.39,-4.87]$};
\node[anchor=south west,font=\scriptsize\bfseries,text=ink] at (axis cs:22,4.55) {\AdS$-$\ObsS{} [95\%]};
\end{axis}
\end{tikzpicture}
\caption{\textbf{Exploratory study: laboratory access lowers mean regret for every deployment.} Grey bars span \Obs{} to \Ad; diamonds mark \Fx. Right: post hoc, descriptive world-bootstrap intervals for \AdS$-$\ObsS. Worlds were previously examined, so these results motivated, rather than confirm, the prospective study.}
\label{fig:exploratory}
\end{figure}

In the exploratory study, all 576 decisions across Sol, Terra, Luna, and the supplementary GPT-5.5 deployment were scored, and each mean uses 48 world--direction units (Table~\ref{tab:exploratory}, Figure~\ref{fig:exploratory}). Laboratory access lowers mean regret for all four deployments. Post hoc world-bootstrap intervals for \AdS$-$\ObsS{} exclude zero for Sol, Terra, and GPT-5.5; Luna's interval includes zero. These results motivated the independent prospective evaluation above.

\begin{table}[t]
\centering
\small
\setlength{\tabcolsep}{4.5pt}
\begin{tabular}{lccccc}
\toprule
Deployment & \Obs & \Ad & \Fx & \AdS$-$\ObsS{} [95\% int.] & \AdS$-$\FxS{} [95\% int.]\\
\midrule
Sol & 6.121 {\scriptsize(37)} & 0.764 {\scriptsize(42)} & 1.091 {\scriptsize(46)} & $-5.357\;[-8.951,-2.195]$ & $-0.327\;[-1.925,\phantom{-}0.786]$\\
Terra & 19.902 {\scriptsize(18)} & 3.550 {\scriptsize(40)} & 4.441 {\scriptsize(41)} & $-16.352\;[-22.760,-9.686]$ & $-0.891\;[-5.934,\phantom{-}4.422]$\\
Luna & 7.676 {\scriptsize(36)} & 4.551 {\scriptsize(39)} & 3.287 {\scriptsize(43)} & $-3.125\;[-8.181,\phantom{-}2.226]$ & $\phantom{-}1.264\;[-3.186,\phantom{-}5.660]$\\
GPT-5.5 (supp.) & 10.017 {\scriptsize(31)} & 1.575 {\scriptsize(45)} & 3.160 {\scriptsize(44)} & $-8.441\;[-12.393,-4.872]$ & $-1.585\;[-4.236,\phantom{-}0.386]$\\
\bottomrule
\end{tabular}
\caption{\textbf{Exploratory mean target regret} on 24 previously examined worlds (48 units per cell); parentheses give optimal-action counts out of 48. Intervals are descriptive and were specified after observing outcomes. Terra combines two recorded configurations (Appendix~\ref{app:details}). GPT-5.5 is supplementary; no pooled effect is estimated.}
\label{tab:exploratory}
\end{table}

\subsection{How much decision information does each laboratory retain?}
A simulator can preserve the best action without reproducing every aspect of its utility. To separate these properties, we use selectors that observe the complete laboratory utility of both interventions, choose the higher one, and commit before independent evaluation on the adaptive target. Table~\ref{tab:selectors} shows that both laboratories preserve the target-optimal choice in nearly all 48 exploratory units. The selectors receive both full laboratory utilities, whereas the LLM agents receive public experiment summaries; this diagnostic therefore measures the decision information a laboratory retains, not a comparison under equal information.

\begin{table}[t]
\centering\small
\begin{tabular}{lcc}
\toprule
Laboratory utility selector & Optimal target actions & Mean target regret\\
\midrule
Fixed-mix laboratory $\rightarrow$ adaptive target & 47/48 & 0.3802\\
Adaptive laboratory $\rightarrow$ adaptive target & 46/48 & 0.5895\\
\bottomrule
\end{tabular}
\caption{\textbf{Exploratory action-ranking diagnostic.} Each selector observes the complete laboratory utilities of both interventions and commits before independent target evaluation. Nonzero target regret reflects the remaining transfer errors.}
\label{tab:selectors}
\end{table}

\section{Discussion}
\label{sec:discussion}

\textbf{The value of an experiment is the decision it changes.} The task links an agent's research process to a consequence: avoided intervention loss. Writing $\bar R_b$ for mean regret under an experimental budget of $b$ calls, the gross value of experimentation is
\begin{equation}
V_b=\bar R_0-\bar R_b,
\end{equation}
the loss avoided relative to observation alone. For Luna on new worlds, $V_8=5.092$ with the adaptive laboratory. This quantity makes experimental usefulness comparable across agents, budgets, and laboratories on a common scale, and a resource-efficiency analysis could weigh it against measured acquisition and inference costs.

\textbf{Which simulator details matter?} For an agent, the relevant question is whether an evolving strategy mix improves the final choice between deleveraging and a liquidity backstop. In our data it does not demonstrably do so: the prospective \AdS$-$\FxS{} interval admits either sign, and complete-utility selectors from both laboratories recover nearly all target-optimal actions. This suggests a concrete evaluation practice: before treating added dynamics as necessary for a task, test whether they change the intervention ranking or reduce target regret~\citep{grimm2020value,lambert2020mismatch}. Because the simulator's mechanisms are switchable by stage, the same test can be applied to each stage. Runtime savings and transfer to real markets remain separate empirical questions.

\textbf{Limitations.} The evidence covers one synthetic crash construction and two interventions, each with a single fixed implementation (one deleveraging sale; one backstop size). The fundamental value is constant, maker depth falls on a fixed schedule at the shock, the evolving strategy mix has two candidates (one of them signal-independent), and the shadow-trade horizon is one step. New constructors and independent target seeds support transfer \emph{within} this construction, not to other constructions or to historical markets. With 24 worlds, the primary effect is clear in direction but modest in two-sided strength (exact sign-flip $p=0.075$). Incomplete Sol and Terra records limit prospective cross-deployment conclusions, and the placebo comparison leaves the contribution of world-specific evidence relative to tool interaction and generic crash information uncertain. The positive primary result therefore concerns the full laboratory-access condition for Luna within this construction. Finally, regret is measured in simulator utility units and is not calibrated to monetary outcomes.

\textbf{Next measurements.} The most discriminating next comparison is a strong fixed acquisition schedule with the same observations and decision rule. It would test whether an agent \emph{selects} experiments well, beyond benefiting from access to informative packets.

\section{Conclusion}
CrashGym measures the decision value of experimentation in a deterministic, replayable market laboratory. Independent target realizations connect acquired evidence to consequential intervention choices and make regret exact for the realized target repeats, and staged amplification mechanisms make simulator detail testable by its contribution to a downstream choice. Prospective results support a laboratory-access benefit for Luna, and a separate exploratory study shows the same descriptive direction across four deployments. The resulting question for financial agents is concrete: which experiments, observations, and simulator details are worth acquiring for the decision at hand?

\begingroup
\small
\setlength{\bibsep}{2pt plus 1pt}

\endgroup

\appendix
\section{Simulator parameters}
\label{app:params}

\begin{table}[h]
\centering\small
\begin{tabularx}{\linewidth}{@{}l>{\raggedright\arraybackslash}X@{}}
\toprule
Parameter & Value\\
\midrule
\multicolumn{2}{@{}l}{\emph{Fixed across worlds}}\\
Horizon; shock step; intervention step; exposure step & $T=96$; $t_0=8$; step 4; step 1\\
Tick size; fee; cost scale $\lambda$ & 1 tick; 1 per lot per counterparty; 100\\
Fundamental value $v$ & 10{,}000 ticks, constant\\
Risk limits & position $10^6$ lots; notional $\le 20\times$ initial cash\\
Marketable-order band & 100 ticks through the best opposite quote\\
Maker ladder & 40 levels per side; half-spread 2 ticks; $\bar Q$ lots/level at step 0 and from the shock on, 400 lots/level in steps 1--7; stress floor 25\%\\
Noise flow & 8 orders/step of 4 lots\\
Fundamental and trend traders & 4 lots per tick of mispricing; 4 lots per half-tick of $e_t=\operatorname{trunc}((3e_{t-1}+r_{t-1})/4)$\\
Persistent impact & $40x/D_0$ ticks, $D_0=10\bar Q$ (top-10 step-0 bid depth)\\
Adaptive participant & initial shares $(\tfrac12,\tfrac12)$; \textsc{Fundamental} capped at 80 lots; \textsc{Trend} pair $(-160,+80)$; $\eta=10^{-3}$\\
Interventions & deleveraging sells $\lceil|q|/2\rceil$ (+2-lot account purchase); backstop: 1-lot purchase, 40-lot bid at best bid each step\\
\midrule
\multicolumn{2}{@{}l}{\emph{Sampled per world}}\\
Strata & neither / liquidation only / withdrawal only / both; 6 worlds each\\
Liquidation strength $\alpha$; withdrawal strength $\beta$ & uniform on $[0.05,1]$; inactive one set to 0\\
Shock size; leveraged exposure & $\{160,180,220,240\}$ lots sold; $\{160,\dots,240\}$ lots bought (step 20)\\
Leveraged cash & $\{84\text{k},\dots,124\text{k}\}$ (step 10k), raised to cover the $20\times$ limit\\
Maintenance margin $\mu$ & $\{3\%,\dots,7\%\}$ (step 1\%)\\
Maker depth $\bar Q$; stress scale $\kappa$ & $\{24,\dots,56\}$ lots (step 8); $\{24,\dots,56\}$ half-ticks (step 8)\\
Backstop level size$^{\dagger}$ & $\{24,\dots,56\}$ lots (step 8); not used by this construction\\
\bottomrule
\end{tabularx}
\caption{\textbf{Parameters of the crash construction} used in both studies. $^{\dagger}$Sampled and included in the constructor hash (hence in duplicate screening), but the backstop size is fixed at 40 lots in this construction.}
\label{tab:params}
\end{table}

\section{Protocol and evidence details}
\label{app:details}

\textbf{Models.} Sol, Terra, and Luna are the three served variants of ChatGPT~5.6.4 (\texttt{gpt-5.6-sol}, \texttt{gpt-5.6-terra}, \texttt{gpt-5.6-luna}); Sol was reached through a separate API gateway. Every request records the exact requested model identifier, and the analysis rejects any lane with mixed identifiers. GPT-5.5 is supplementary in the exploratory study and absent from the prospective confirmation. The exploratory Terra results combine 143 decisions under one recorded configuration and one \Obs{} decision under another; equivalence of the underlying checkpoints was not independently verified.

\textbf{Interaction limits.} Besides the turn and correction limits in Section~\ref{sec:protocol}, an episode terminates without an action if a response cannot be parsed, a single turn exceeds 10,000 output tokens, or the episode exceeds 320,000 output tokens. Provider-side reasoning content is not returned to the agent in later turns.

\textbf{Sampling and uncertainty.} The prospective population contains six worlds in each of four strata. The two seed cohorts have eight repeats each, giving 384 unique seed commitments. The analysis uses 10,000 stratum-preserving whole-world bootstrap draws with PCG64 seed 20260908 and linear 2.5/97.5 percentiles. Intervals are nominal and deployment-specific, with no joint or multiplicity-adjusted success claim. The exploratory intervals come from a post hoc analysis and are descriptive.

\textbf{Missing outcomes.} The prospective study ended with no pending decisions: Luna 192 scored; Sol 88 scored, 13 empty, and 91 infrastructure-blocked; Terra 82 scored and 110 empty. The infrastructure stop followed a permanent request error. Missing actions receive no numerical regret. The 110 Terra failures are provider responses that could not be parsed as a message; the 13 Sol failures are episodes ending after three consecutive responses without exactly one tool call. Table~\ref{tab:census} shows that Terra's failures concentrate in the three tool conditions (94 of 110), so missingness depends on condition and complete pairs are not a random subsample. The full-denominator primary comparisons for Sol and Terra are therefore not replaced by complete-case estimates. For transparency only, the complete-pair mean \AdS$-$\ObsS{} differences are $-8.221$ for Sol (20 pairs) and $-6.583$ for Terra (16 pairs); these are not estimates of the planned contrasts. Separate post hoc recovery attempts are outside the reported confirmation results.

\begin{table}[h]
\centering
\small
\begin{tabular}{llcccc}
\toprule
Deployment & Condition & Scored & Empty & Blocked & Mean calls (range)\\
\midrule
Luna & \Obs & 48 & 0 & 0 & ---\\
 & \Ad & 48 & 0 & 0 & 4.71 (4--7)\\
 & \Fx & 48 & 0 & 0 & 4.52 (2--6)\\
 & \Pl & 48 & 0 & 0 & 4.25 (3--7)\\
\midrule
Sol & \Obs & 25 & 1 & 22 & ---\\
 & \Ad & 22 & 4 & 22 & 4.45 (3--6)\\
 & \Fx & 22 & 3 & 23 & 4.36 (3--6)\\
 & \Pl & 19 & 5 & 24 & 4.42 (3--8)\\
\midrule
Terra & \Obs & 32 & 16 & 0 & ---\\
 & \Ad & 17 & 31 & 0 & 4.35 (3--6)\\
 & \Fx & 18 & 30 & 0 & 4.56 (4--6)\\
 & \Pl & 15 & 33 & 0 & 4.13 (3--6)\\
\bottomrule
\end{tabular}
\caption{\textbf{Prospective census and laboratory use} by deployment and condition (48 planned decisions per row). Experiment calls are counted over scored episodes, with a budget of eight.}
\label{tab:census}
\end{table}

\textbf{Replay.} An independent replay reproduces all 362 available prospective action scores and preserves the 214 missing outcomes. A separate replay verifies the complete exploratory decision transcripts and all 576 scores. \par\textbf{Code and data availability.} Source code, configurations, transcripts, and replay receipts are retained by the authors; a public artifact package has not yet been released.

\textbf{Binary companion tests.} The prespecified direction-specific tests of Luna's optimal-action outcomes yield $p=0.5811$ and $p=0.3877$. Neither meets the directional criterion, despite the supported primary regret reduction.

\textbf{Metric interpretation.} Utility combines the reference-drawdown objective with implementation costs, and the score compares the chosen intervention with the best feasible one on the same target repeats. Experiment packets use intervention-specific controls, so their paired differences need not directly rank full target utilities. Successful use of these packets is evidence of decision performance, not proof that an agent recovered the market's latent causal structure.

\end{document}